\documentclass[conference]{IEEEtran}
\IEEEoverridecommandlockouts

\usepackage{amsmath,amssymb,amsfonts,amsthm}

\usepackage{graphicx}
\usepackage{textcomp}
\usepackage{booktabs} 
\usepackage{siunitx}  
\usepackage{tikz}
\usetikzlibrary{arrows, shapes, positioning}
\usepackage{bm}
\usepackage{color}

\def\BibTeX{{\rm B\kern-.05em{\sc i\kern-.025em b}\kern-.08em
    T\kern-.1667em\lower.7ex\hbox{E}\kern-.125emX}}
\usepackage{verbatim}
\usepackage{dsfont}
\usepackage{subcaption}
\usepackage{mathtools}
\usepackage{multirow}

\newcommand\figref{Figure~\ref}
\newtheorem{theorem}{Theorem}

\newtheorem{remark}{Remark}

\usepackage{array}
\makeatletter
\newcommand{\ostar}{\mathbin{\mathpalette\make@circled\star}}
\newcommand{\make@circled}[2]{%
  \ooalign{$\m@th#1\smallbigcirc{#1}$\cr\hidewidth$\m@th#1#2$\hidewidth\cr}%
}
\newcommand{\smallbigcirc}[1]{%
  \vcenter{\hbox{\scalebox{0.77778}{$\m@th#1\bigcirc$}}}%
}
\makeatother

\usepackage{kotex}
\usepackage{bbm}
\usepackage{mathtools}
\usepackage{multirow}
\usepackage{textcomp}
\usepackage{cite}
\usepackage{amsmath,amssymb,amsfonts}
\usepackage{algorithm}
\usepackage[noend]{algpseudocode}
\usepackage{graphicx}
\usepackage{textcomp}
\usepackage{xcolor}
\def\BibTeX{{\rm B\kern-.05em{\sc i\kern-.025em b}\kern-.08em
    T\kern-.1667em\lower.7ex\hbox{E}\kern-.125emX}}
\begin{document}
\title{\fontsize{23}{28}\selectfont Uncertainty-Aware Hybrid Inference with \\On-Device Small and Remote Large Language Models}

\author{
\IEEEauthorblockN{Seungeun Oh$^{1*}$, Jinhyuk Kim$^{1*}$, 
Jihong Park$^{2}$, 
Seung-Woo Ko$^{3}$,
Tony Q. S. Quek$^{2}$,
and Seong-Lyun Kim$^{1}$}

\IEEEauthorblockA{\centering $^{1}$School of EEE, Yonsei University, \{seoh, jh.kim, slkim\}@ramo.yonsei.ac.kr}

\IEEEauthorblockA{\centering $^{2}$ISTD Pillar, Singapore University of Technology and Design, \{jihong\_park, tonyquek\}@sutd.edu.sg}

\IEEEauthorblockA{\centering $^{3}$Department of SME, Inha University, swko@inha.ac.kr}
}

\maketitle 
\def\thefootnote{*}\footnotetext{Equal contribution}\def\thefootnote{\arabic{footnote}}

\begin{abstract}
This paper studies a \textit{hybrid language model (HLM)} architecture that integrates a \textit{small language model (SLM)} operating on a mobile device with a \textit{large language model (LLM)} hosted at the base station (BS) of a wireless network. The HLM token generation process follows the speculative inference principle: the SLM's vocabulary distribution is uploaded to the LLM, which either accepts or rejects it, with rejected tokens being resampled by the LLM. While this approach ensures alignment between the vocabulary distributions of the SLM and LLM, it suffers from low token throughput due to uplink transmission and the computation costs of running both language models. To address this, we propose a novel HLM structure coined \textit{Uncertainty-aware opportunistic HLM (U-HLM)}, wherein the SLM locally measures its output uncertainty and skips both uplink transmissions and LLM operations for tokens that are likely to be accepted. This opportunistic skipping is enabled by our empirical finding of a linear correlation between the SLM's uncertainty and the LLM's rejection probability. We analytically derive the uncertainty threshold and evaluate its expected risk of rejection. Simulations show that U-HLM reduces uplink transmissions and LLM computations by 45.93\%, while achieving up to 97.54\% of the LLM's inference accuracy and 2.54$\times$ faster token throughput than HLM without skipping.
\end{abstract}

\begin{IEEEkeywords}
Large language model (LLM), on-device LLM, speculative inference, uncertainty, opportunistic transmission.
\end{IEEEkeywords}

\vspace{-7pt}
\section{Introduction}
Recent advancements in large language models (LLMs) have shown that the scaling model and training data sizes according to power laws enable more emergent capabilities \cite{hoffmann2022training}. This scalability has made LLMs effective across various applications, including AI agents, multimodal reasoning, and even human-level intelligent robots \cite{yang2024harnessing}. However, after training completes, LLM inference requires huge and high bandwidth memory, which is not feasible in mobile devices, limiting their adoption mostly within powerful servers. Recently, on-device langauge models, also known as small language models (SLMs), have been actively developed through techniques such as pruning pre-trained LLM parameters, quantizing model parameters and activations, and distilling knowledge from LLMs into SLMs \cite{ma2023llm, li2023loftq, zhou2023distillspec}. However, while these techniques enable significantly lighter models suitable for deployment on mobile devices, they often result in trade-offs, including reduced accuracy and diminished capabilities compared to their large-scale counterparts.

Under this trade-off, a promising approach for enabling LLM inference over wireless networks is the \textit{hybrid language model (HLM)} \cite{hao2024hybrid}, which leverages \textit{speculative inference} \cite{leviathan2023fast} to facilitate the joint utilization of an LLM and an SLM, deployed on a server at the base station (BS) and a mobile device, respectively. In HLM, given a vocabulary—a complete set of basic units called tokens—the SLM processes the input token sequence to generate a probability distribution over the vocabulary, referred to as the vocabulary distribution, followed by sampling draft tokens. Similarly, the LLM computes its own vocabulary distribution, and the probability corresponding to the draft token in each vocabulary distribution is compared to determine the draft token's acceptance or rejection. If the draft token is rejected, resampling is performed to produce the target token, which becomes the response token of the HLM. This method can replicate the vocabulary distribution achieved in traditional LLM inference, where the LLM generates response tokens directly from the input token sequence, while efficiently leveraging the distributed computing resources and memory of both the mobile device and the server.

\begin{figure}[t]
\centering
\includegraphics[width=0.49\textwidth]{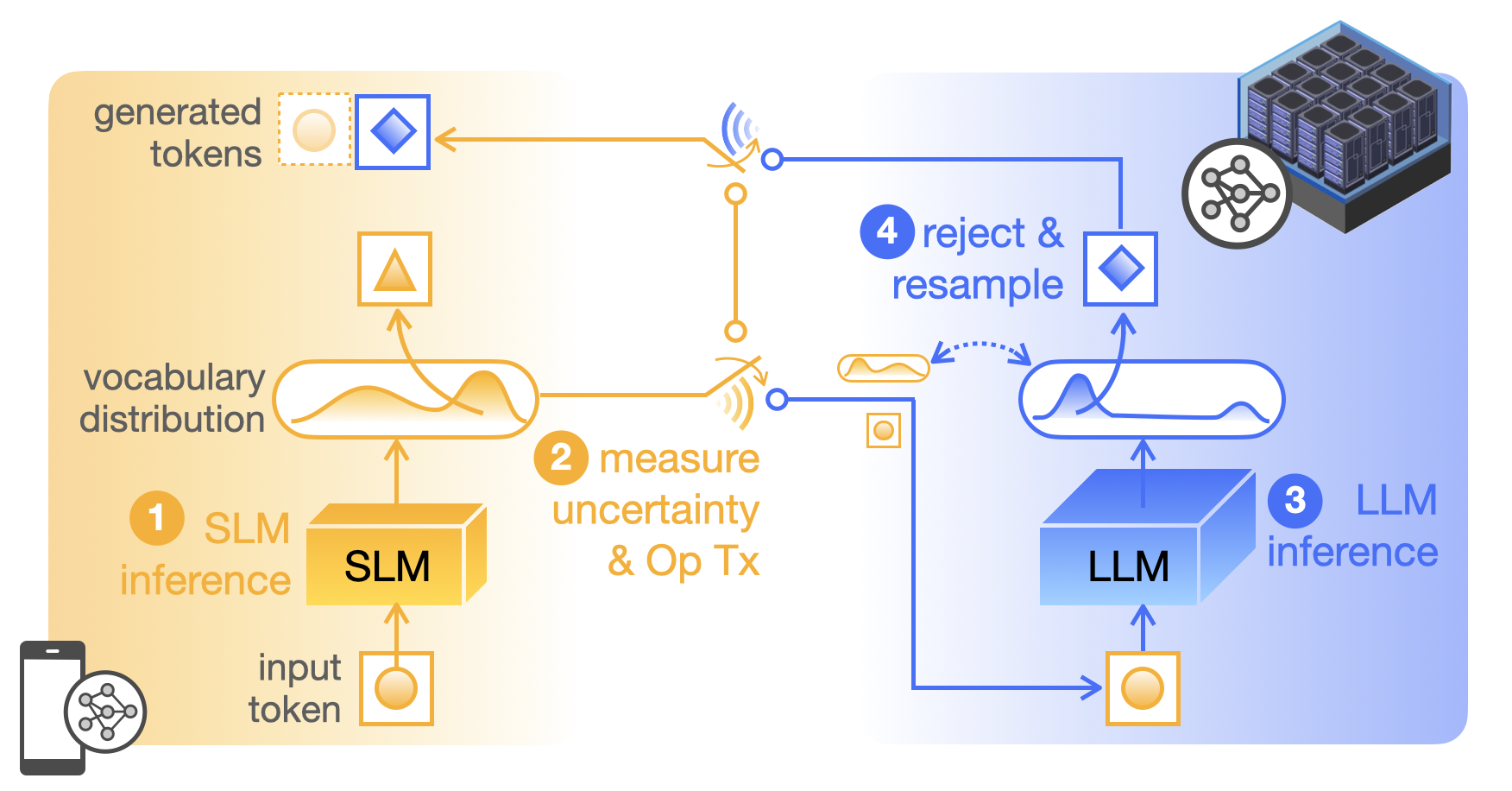}
\caption{Schematic illustration of the proposed U-HLM over a wireless network consisting of a single device-server.}
\vspace{-15pt}
\label{fig:system_model}
\end{figure}
\vspace{-1pt}
Despite these potential benefits, HLM faces a significant challenge in achieving high token throughput. To generate a single token, the device must upload a vocabulary distribution, with a payload size proportional to the dimension of the entire vocabulary (i.e., $500$ kbit for a vocabulary size of $32,000$). Additionally, the process involves computational overhead from both the SLM and LLM, requiring, for example, 24.6 ms and 104.6 ms, respectively, in our experimental setup, as detailed in Section \ref{Sec: 5}. This low throughput not only increases latency but also significantly degrades user experience in LLM serving systems, further restricting the practicality of HLM for real-time or near-real-time applications.

To this end, this paper seeks to find a solution for the HLM framework that improves token throughput while minimizing the degradation of inference accuracy. The approach builds on a straightforward yet effective strategy: skip uplink transmissions as well as LLM computations for tokens that are likely to be accepted. However, implementing this is challenging, as the SLM must independently predict whether a token will be accepted or rejected without access to the LLM's vocabulary distribution. To enable this capability, we explore a method in which the SLM leverages \textit{uncertainty}, a measure of the model's self-assessed confidence in its outputs, to predict the LLM's rejection probability. Among the various approaches we evaluate for measuring uncertainty in LLMs, our experiments reveal that uncertainty measured through temperature perturbation exhibits a strong linear correlation with the rejection probability in the HLM framework. Building on this observation, we propose a novel HLM framework called \textit{Uncertainty-aware opportunistic HLM (U-HLM)}. As illustrated in \figref{fig:system_model}, U-HLM enables the device to selectively transmit the SLM's vocabulary distribution only for tokens with uncertainty exceeding a predefined threshold.

Beyond this, we design the uncertainty threshold for U-HLM, aiming to skip tokens that are not only immediately accepted but also those likely to be probabilistically accepted. We also derive a theoretical upper bound on the loss of inference accuracy in terms of expected rejection risk associated with the uncertainty threshold and demonstrate that it is negligible when empirical values are applied. Through experiments, we show that U-HLM achieves inference accuracy comparable to that of an LLM while significantly improving token throughput, enabled by its highly accurate and frequent skipping. This advantage becomes particularly pronounced under poor channel conditions, highlighting U-HLM's effectiveness in resource-constrained environments.
 \vspace{-5pt}
\section{System Model}  \vspace{-3pt}
In this study, we consider a network architecture consisting of a single device and a base station (BS) with a powerful computational server. We assume that an SLM is deployed on the device, while an LLM is deployed on the server, following the architecture in \cite{hao2024hybrid}. In this network, both the device and the server operate with \textit{tokens} as their basic units, where the \textit{vocabulary} $\mathcal{V}$ represents the full set of possible tokens. We suppose that this vocabulary $\mathcal{V}$ is shared by SLM and LLM, and define \textit{initial prompt sequence} as $\mathbf{s}$, serving as input in the initial phase of inference.
\vspace{-5pt}
\subsection{Token Generation of Hybrid Lauguage Model} \label{hybrid_infer}
The core behavior of hybrid language model (HLM) is that the SLM on the device generates a \textit{draft token}, which is then either accepted or rejected and resampled by the LLM on the server. The detailed behaviour of this is as follows.

\noindent \textbf{Step 1) SLM's Generation.} \quad 
In the \(t\)-th round, the \textit{input token sequence}, \(\mathbf{s}(t-1)\), is defined as the combination of the initial prompt sequence \(\mathbf{s}\) and the cumulative sequence \(\mathbf{r}(t-1)\) of \textit{response tokens}. Here, \(\mathbf{r}(t-1)\) comprises tokens generated from the \(1\)-st to the \((t-1)\)-th round, with one token produced per round (specifically, the response token \(r(t-1)\) generated in the \((t-1)\)-th round). Formally, we define \(\mathbf{s}(t-1) \coloneqq \mathbf{s} \oplus \mathbf{r}(t-1)\), where \(\oplus\) denotes the concatenation operator.

The device's SLM processes the input token sequence to produce a logit vector $\mathbf{z}(t) = [z_{1}(t),z_{2}(t),... .,z_{|\mathcal{V}|}(t)]^\top$. This vector is then normalized into a vocabulary distribution $\mathbf{x}(t) = [x_{1}(t),x_{2}(t),...,x_{|\mathcal{V}|}(t)]^\top$, where each element is defined by:
\begin{align}
    x_{v}(t) = \frac{\exp(z_{v}(t))}{\sum_{i=1}^{|\mathcal{V}|} \exp(z_{i}(t))}, \quad \forall v \in \mathcal{V}. \label{eq: local_process}
\end{align}
Next, the draft token $d$ is sampled from the SLM's vocabulary distribution, $d\sim\mathbf{x}(t)$. The device then determines whether the SLM's draft token $d$ should serve as the response token $r(t)$ for the $t$-th round. If the decision is to proceed with the draft token, \textbf{Step 2)} of the $t$-th round is skipped.

\noindent \textbf{Step 2) LLM's Verification.} \quad This step employs speculative inference \cite{leviathan2023fast}, ensuring that the vocabulary distribution of the response token in HLM inference matches that of an LLM inference. This equivalence is rooted in the principles of the Metropolis-Hastings algorithm \cite{chib1995understanding}.

The LLM on the server processes \(\mathbf{r}(t-1)\) in a similar manner to generate a logit vector, followed by the LLM's vocabulary distribution \(\mathbf{y}(t) = [y_{1}(t), y_{2}(t), \ldots, y_{|\mathcal{V}|}(t)]^\top\). Here, for the draft token \(d\), \(x_{d}(t)\) and \(y_{d}(t)\) are referred to as the draft and target probabilities, respectively. If \(x_{d}(t) \leq y_{d}(t)\), the draft token is immediately accepted as the response token for the \(t\)-th round (immediate acceptance). Otherwise, it is rejected with probability \(1 - y_{d}(t)/x_{d}(t)\), or accepted otherwise (probabilistic acceptance/rejection). When a rejection occurs, the LLM calculates a normalized distribution based on the difference between the vocabulary distributions of the LLM and the SLM, where the $v$-th probability is given by:
\begin{equation}
P_{v}(t) = \frac{\max(y_{v}(t) - x_{v}(t), 0)}{\sum_{i=1}^{|\mathcal{V}|} \max(y_{i}(t) - x_{i}(t), 0)}, \quad \forall v \in \mathcal{V}.
\label{eq: resample}
\end{equation}
This is followed by resampling the LLM's target token \(d^\ast \sim \text{norm}\left(\max(\mathbf{y}(t) - \mathbf{x}(t), 0)\right)\). In this case, the response token \(r(t)\) is replaced with the target token \(d^\ast\).

 \begin{figure*}[t]
 \centering
\includegraphics[width=0.99\linewidth]{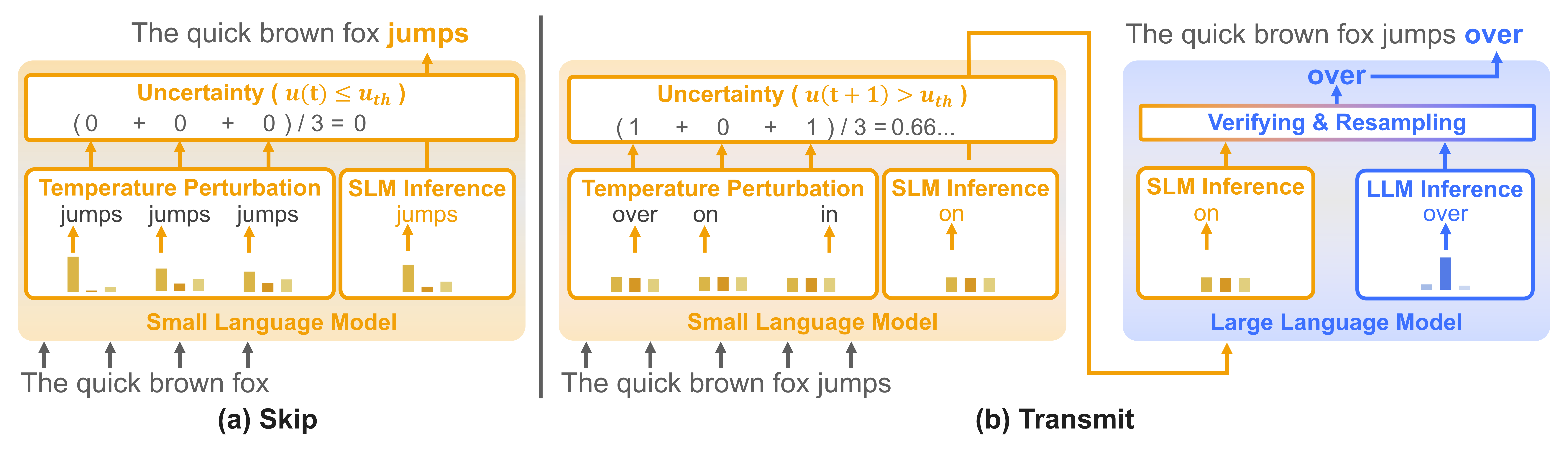}
\vspace{-10pt}
\caption{Detailed process of generating response tokens in U-HLM with \( u_{\text{th}} = 0.5 \): The input token sequence is processed by the SLM in two ways—one for generating the SLM's draft token and another for temperature perturbation. Tokens sampled from the temperature perturbation are compared with the draft token, returning 0 if they match and 1 otherwise, with the average used to compute the uncertainty \( u(t) \). (a) When \( u(t) \leq u_{\text{th}} \), uplink transmission and LLM operation are skipped. (b) When \( u(t) > u_{\text{th}} \), the process continues with the LLM's verification and resampling.}
 \label{fig:3 Proposed}
 \end{figure*}

\noindent \textbf{Step 3) SLM's Concatenation.} \quad For a selected $r(t)$, the device generates the response token sequence $\mathbf{r}(t)$ for the $t$-th round by concatenating $r(t)$ to the existing sequence $\mathbf{r}(t-1)$ as follows: $\mathbf{r}(t)\coloneqq \mathbf{r}(t-1)\oplus r(t) $. Following this update, the process proceeds to \textbf{Step 1)} of the $(t+1)$-th round, which implies LLM's autoregressive generation property, where the response token from the current round is incorporated into the input token sequence for the next round.

This iterative generation continues until reaching a stopping condition: either the total number of response tokens reaches a specified maximum length, $|\mathbf{r}(t)|=r_{\text{max}}$, or the language model selects an End-of-Sentence (EOS) token, $r(t)=\text{EOS}$, indicating the completion of response generation for $\mathbf{s}$. Once terminated, the device resets with a new initial prompt sequence $\mathbf{s}'$, starting an independent token generation.
\subsection{Wireless Communication}
The operation involves uplink and downlink transmissions between the device and the BS. In the $t$-th round, after the device generates the draft token $d$ in the SLM and decides to send it, the following information is transmitted to the BS over the uplink: 1) the index \(v\) of the draft token selected from the vocabulary \(\mathcal{V}\), and 2) the SLM's vocabulary distribution \(\mathbf{x}(t)\). The BS then returns the index of the target token $d^\ast$ to the device via the downlink if resampling is performed.

\noindent \textbf{Channel Model.} \quad Since the size of the index is negligible compared to the vocabulary distribution, we simplify the analysis by considering only the uplink transmission latency for the vocabulary distribution. The uplink payload size $B$ is calculated as: $B=|\mathcal{V}|\cdot b_{\text{prob}}\text{ bits}$, where $b_{\text{prob}} = 32$ bits for full precision and $16$ bits for half precision. We consider a block fading channel such that channel gains are constant over one round and may change over different rounds. Using Shannon's formula, we can express the uplink transmission time $\tau(t)$ for the $t$-th round as:
\begin{equation}
  \tau(t) = \frac{B}{W \log_2{(1+\text{SNR}(t))}}, 
  \label{SNR}
\end{equation}
where $W$ represents the uplink channel bandwidth. Here, the received signal-to-noise ratio (SNR) depends on the transmission power $p$,  the distance $\rho$ between the device and the BS, the path loss exponent $\alpha$, and the noise power $N$ under Rayleigh fading conditions.

\noindent \textbf{Token Throughput.} \quad Token throughput is defined as the number of tokens generated per unit time, accounting for both communication and computation times. To formalize this, we introduce a binary variable \(\delta(t) \in \{0,1\}\), which indicates whether uplink transmissions and LLM computation are performed (\(\delta(t) = 1\)) or skipped (\(\delta(t) = 0\)) in the \(t\)-th round. Let \(\tau_{\text{SLM}}\) and \(\tau_{\text{LLM}}\) denote the computation times required by the SLM and LLM, respectively, to generate a single token. For the \(t\)-th round of token generation, the token throughput \(T(t)\) is given by:
\begin{equation}
T(t) \text{ (in token/sec)} =
\begin{cases}
\frac{1}{\tau_{\text{SLM}} + (\tau(t) + \tau_{\text{LLM}})}, & \text{if $\delta(t) = 1$,} \\
\frac{1}{\tau_{\text{SLM}}}, & \text{if $\delta(t) = 0$.}
\end{cases}
\label{eq:spec_latency}
\end{equation}

This HLM structure effectively utilizes distributed network resources and achieves high inference accuracy but suffers from low token throughput due to the communication and computation required for generating each token. A straightforward way to mitigate this issue without compromising inference accuracy is to skip transmission and computation for tokens that are likely to be accepted if sent to the server. However, implementing such a skipping mechanism in the current HLM framework requires knowledge of both draft and target probabilities (denoted by $x_{d}(t)$ and $y_{d}(t)$, respectively). While the draft probability is available on the device, the target probability is only accessible from the server. To make this approach feasible, an additional mechanism is needed to predict token acceptance or rejection without relying on target probability information from the server.
\section{Uncertainty-Aware Hybrid Language Model for High-Throughput LLM Inference} 
In this section, we conduct a feasibility experiment inspired by the concept of \textit{uncertainty} to investigate whether a device can predict the rejection probability of a token on its own without the server's target probability. Based on this, we propose an \textit{Uncertainty-aware opportunistic Hybrid Language Model (U-HLM)}, in which the SLM measures the uncertainty and opportunistically skips uplink transmissions to the LLM if the estimated uncertainty does not exceed its threshold. Additionally, we design the uncertainty threshold for U-HLM based on the rejection probability threshold that determines when the device intends to skip transmissions. We also derive the expected rejection risk and its upper bound by incorporating the density of uncertainty into the analysis.

\begin{figure}[t]
\centering
\includegraphics[width=0.48\textwidth]{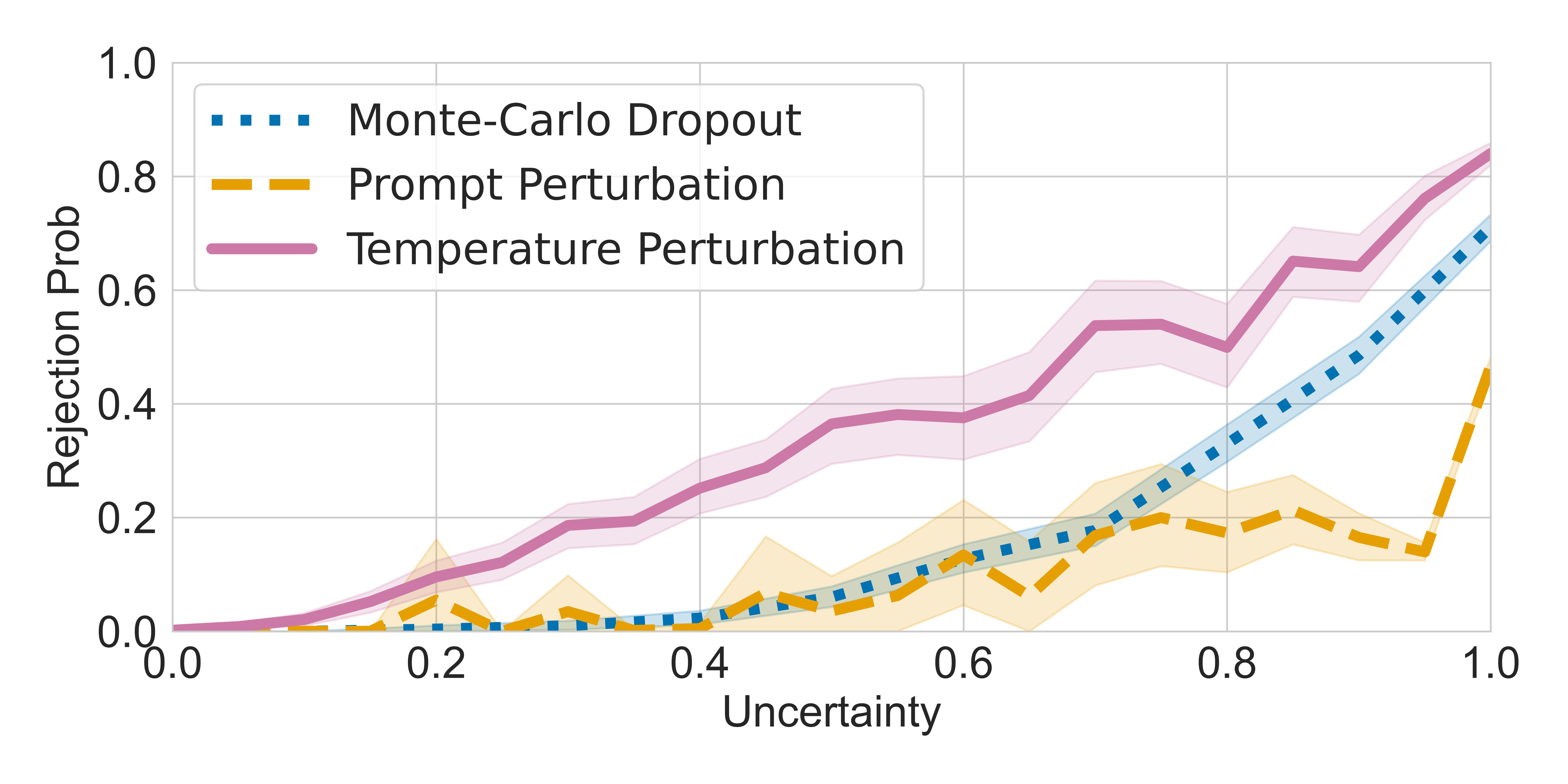}
\vspace{-5pt}
\caption{Curves depicting the relationship between uncertainty and rejection probability for three uncertainty measures. Each curve includes a total of $5,134$ data points, with the line representing the mean and the shaded area indicating the 95\% confidence interval.}
\vspace{-15pt}
\label{fig:uncertainty_comparison}
\end{figure}

\subsection{\small Feasibility of Predicting Rejection Probability Using Uncertainty} \label{section:feasible} 
Our approach is inspired by the concept of \textit{uncertainty}, which quantifies a model's confidence in its output. Specifically, we hypothesize that by measuring the SLM's uncertainty regarding a generated draft token, we can predict the rejection probability of that token in the LLM without requiring the target probability. We conduct the following experiment to explore the feasibility of the above hypothesis. Before proceeding, we define the rejection probability and uncertainty for the draft token $d$ in the $t$-th round of HLM inference as $\beta_d(t) \coloneqq \max\left(1 - \frac{y_d(t)}{x_d(t)}, 0\right)$ and $u(t)$, respectively.

\noindent \textbf{Experiment Setting.} \quad We consider three established techniques: Monte Carlo (MC) dropout \cite{gal2016dropout}, prompt perturbation \cite{huang2023look}, and temperature perturbation \cite{gao2024spuq}. All three methods involve sampling, configured as follows: MC dropout samples draft tokens, each from $20$ models, with each model's dropout probability uniformly distributed between $[0,0.1]$; prompt perturbation samples $20$ paraphrased versions of a text prompt using WordNet \cite{miller1995wordnet}, thereby leading to HLM inference; temperature perturbation involves the temperature sampling, generating $20$ temperature samples from a uniform distribution ranging from $0$ to $\theta_{\text{max}}=2$. The SLM, LLM, and dataset used are the same as those described in Section \ref{Sec: 5}.

\noindent \textbf{Observations.} \quad
\figref{fig:uncertainty_comparison} displays the correlation curves between uncertainty and rejection probability. For all three uncertainty measures, we observe that uncertainty tends to be proportional to rejection probability. Among these, only temperature perturbation exhibited a strong linear relationship between its uncertainty and the rejection probability, while also spanning a wider range of uncertainty. Notably, as will be described in Subsection \ref{3A} with its specific operation, the temperature perturbation can be computed in parallel with the SLM's existing forward propagation for draft generation, thereby avoiding any additional latency. Based on these observations, we state the following:
\vspace{-5pt}
\begin{remark} In the HLM, the uncertainty measured using temperature perturbation demonstrates a linear relationship with the rejection probability, as expressed below:
\begin{equation} \beta_d(t) = a \cdot u(t) + b,  \quad \forall t. \label{eq:uncertainty_rejection2} \end{equation}
\end{remark}
\noindent In our setup, linear regression yields a curve with $a=0.82$ and $b=-0.06$, as shown in \figref{fig:uncertainty_prob_dist} (right).
\vspace{-5pt}
\subsection{Uncertainty-Aware Opportunistic Hybrid Language Model} \label{3A} \vspace{-3pt}
Leveraging this, we design a U-HLM with temperature perturbations, which is detailed below.

\noindent \textbf{U-HLM w. Temperature Perturbation.} \quad In the $t$-th round, the device first samples $K$ temperatures in $[0,\theta_{\text{max}}]$, where we will refer to the $k$-th of them as $\theta_k$. In parallel with the traditional forward propagation of the SLM as described in \textbf{Step 1)}, the input token sequence $\mathbf{s}(t-1)$ along with the given $\theta_k$ passes through the SLM to obtain a vocabulary distribution $\tilde{\mathbf{x}}_{k}(t) = [\tilde{x}_{1,k}(t),\tilde{x}_{2,k}(t),...,\tilde{x}_{|\mathcal{V}|,k}(t)]^\top$, where its $v$-th probability is given by: 
\begin{align}
    \tilde{x}_{v,k}(t) = \frac{\exp(z_{v}(t))/\theta_k}{\sum_{i=1}^{|\mathcal{V}|} \exp(z_{i}(t))/\theta_k}, \quad \forall v \in \mathcal{V}, \label{eq:temp_perturb}
\end{align}
leading to the selection of the $k$-th \textit{sample token}, denoted by $d_k$ ($d_k \sim \tilde{\mathbf{x}}_{k}(t)$). After generating the sample tokens, the device quantifies the uncertainty $u(t)$ for the draft token $d$ as:
\begin{equation} 
    u(t) = \frac{1}{K} \sum_{k=1}^{K} \mathds{1}(d_k \neq d),
    \label{eq:uncertainty} 
\end{equation}
where $\mathds{1}(d_k \neq d)$ is the exact match function, which returns $1$ if $d_k \neq d$ and $0$ otherwise. Consequently, the proposed U-HLM determines whether to skip uplink transmissions based on the uncertainty threshold $u_{\text{th}}$ as follows:
\begin{equation}
\delta(t) =
\begin{cases}
0, & \text{if $u(t)\leq u_{\text{th}}$,} \\
1, & \text{if $u(t)> u_{\text{th}}$.}
\label{eq:u_ops}
\end{cases}
\end{equation}
The remaining behavior follows \textbf{Step 2)} and \textbf{Step 3)} in Subsection \ref{hybrid_infer}, with the exemplary process illustrated in \figref{fig:3 Proposed}. An important consideration in U-HLM is that skipping uplink transmissions can cause a mismatch in the synchronization of the input token sequence for subsequent LLM computations, potentially requiring additional communication for token synchronization between the SLM and LLM; however, this paper neglects it due to the small payload size.
\vspace{-3pt}
\subsection{Guideline for Setting Uncertainty Threshold} 
This subsection outlines the design of an uncertainty threshold for selectively skipping uplink transmissions. Specifically, the device aims to omit uplink transmission for draft tokens whose predicted rejection probability \(\beta_d(t)\) meets the condition \(\beta_d(t) \leq \beta_{\text{th}}\), where \(\beta_{\text{th}}\) represents the rejection probability threshold. Two skipping strategies are considered: risk-averse skipping, which skips only tokens predicted to be immediately accepted (\(\beta_{\text{th}} = 0\)), and risk-prone skipping, which skips both tokens predicted to be immediately accepted and those probabilistically accepted. These strategies involve a trade-off between inference accuracy and token throughput. 
Risk-averse skipping is lossless, ensuring that the vocabulary distribution remains consistent with LLM inference. To formalize this, we extend the definition of rejection probability to any vocabulary token $v$ at time $t$ as: $\beta_v(t) \coloneqq \max\left(1 - \frac{y_v(t)}{x_v(t)}, 0\right)$. Using this, the preserved vocabulary distribution under risk-averse skipping satisfies the unbiasedness requirement, expressed as: \begin{equation} x_v(t)\cdot(1-\beta_v(t)) + \sum_{i=1}^{|\mathcal{V}|}{(x_i(t)\cdot\beta_i(t))}\cdot P_v(t) = y_v(t), \quad \forall v,t, \label{equality} \end{equation}
with the proof of this equality provided in \cite{leviathan2023fast}. However, this approach does not facilitate higher token throughput. In contrast, risk-prone skipping increases token throughput by deliberately relaxing the equality in \eqref{equality}, introducing a bias \(\beta_{\text{th}} > 0\) to $\beta_v(t)$. This adjustment, however, may come at the expense of reduced inference accuracy. In this context, this subsection focuses on risk-prone skipping, aiming to maximize token throughput, derive its uncertainty threshold, and examine the associated risks.

Revisiting the rejection scenario in U-HLM, rejection occurs when the target probability \(y_d(t)\) is less than the draft probability \(x_d(t)\), with a probability of \(1 - y_d(t)/x_d(t)\). For the U-HLM device aiming to skip uplink transmissions, \(x_d(t)\) is deterministic on the device, while \(y_d(t)\) is treated as a random variable. Under these conditions, the rejection probability \(\beta_d(t)\) can be reformulated as follows, incorporating the stochastic nature of \(y_d(t)\):
\begin{equation}
\beta_d(t)= P(y_d(t) < x_d(t)) \cdot \textsf{E}_{y_d(t)}\left[ \frac{y_d(t)}{x_d(t)} \mid y_d(t) < x_d(t) \right]. \label{eq:10}
\end{equation}
Ideally, the uncertainty threshold should be adjusted individually for each $t$-th round. However, for simplicity, we adopt a fixed threshold. To determine this threshold, we derive an uncertainty bound under the assumption that uncertainty and rejection probability are independently and identically distributed (i.i.d.) across rounds:
\vspace{-3pt}
\begin{theorem} \label{theorem:combined} 
Under the i.i.d. assumption, with $u \coloneqq u(t)$ and $\beta \coloneqq \beta(t)$ for any $t$, the uncertainty threshold $u_{\text{th}}$ in U-HLM is given by:
\begin{equation}
u_{\text{th}} = \frac{\Delta-b}{a},
\label{Eq:skip}
\end{equation}
where $\Delta=P(y_d < x_d)$ represents the probability that a draft token $d$ is either probabilistically accepted or rejected.

\noindent Defining $R$ as the expected rejection risk, where $f(u)$ denotes the probability density function (PDF) of uncertainty:
\begin{equation}
R = \int_{u=-\frac{b}{a}}^{\frac{\Delta - b}{a}} (au+b) \cdot f(u) \, du,
\end{equation}
we can calculate the upper bound of $R$ as follows:
\begin{align}
R \leq \sqrt{\int_{u=-\frac{b}{a}}^{\frac{\Delta - b}{a}} |au + b|^2 \, du} 
\cdot 
\sqrt{\int_{u=-\frac{b}{a}}^{\frac{\Delta - b}{a}} |f(u)|^2 \, du}\\
= \frac{\Delta^{3/2}}{\sqrt{3a}}\sqrt{\int_{u=-\frac{b}{a}}^{\frac{\Delta - b}{a}} |f(u)|^2 \, du}.
\label{upperbound}
\end{align}

\begin{proof}
Since $0 \leq \textsf{E}_{y_d}\left[ \frac{y_d}{x_d} \mid y_d < x_d \right] < 1$, it follows that $0 \leq \beta < P(y_d < x_d) = \Delta$ from \eqref{eq:10}. Applying \eqref{eq:uncertainty_rejection2} yields:
\begin{equation}
-\frac{b}{a} \leq u < \underbrace{\frac{\Delta-b}{a}}_{:=u_{\text{th}}},
\label{bounds}
\end{equation}
which concludes the derivation of \eqref{Eq:skip}. The upper bound of $R$ is obtained by applying H{\"o}lder's inequality.
\end{proof}
\end{theorem} \vspace{-5pt}
In \eqref{bounds}, setting the threshold at the upper bound corresponds to risk-prone skipping, while setting it at the lower bound corresponds to risk-averse skipping. In \textbf{Theorem} \ref{theorem:combined}, $R$ quantifies the expected increase in rejection probability when adopting risk-prone skipping instead of risk-averse skipping, reflecting the density of uncertainty. This captures the theoretical impact on U-HLM's inference accuracy, where its loss arises when skipping the uplink for tokens at risk of rejection. This impact, along with the expected rejection risk and its upper bound derived in \textbf{Theorem} \ref{theorem:combined}, is thoroughly validated through the experiments presented in Section \ref{Sec: 5}.

\begin{figure}[t]
\centering
\includegraphics[width=0.48\textwidth]{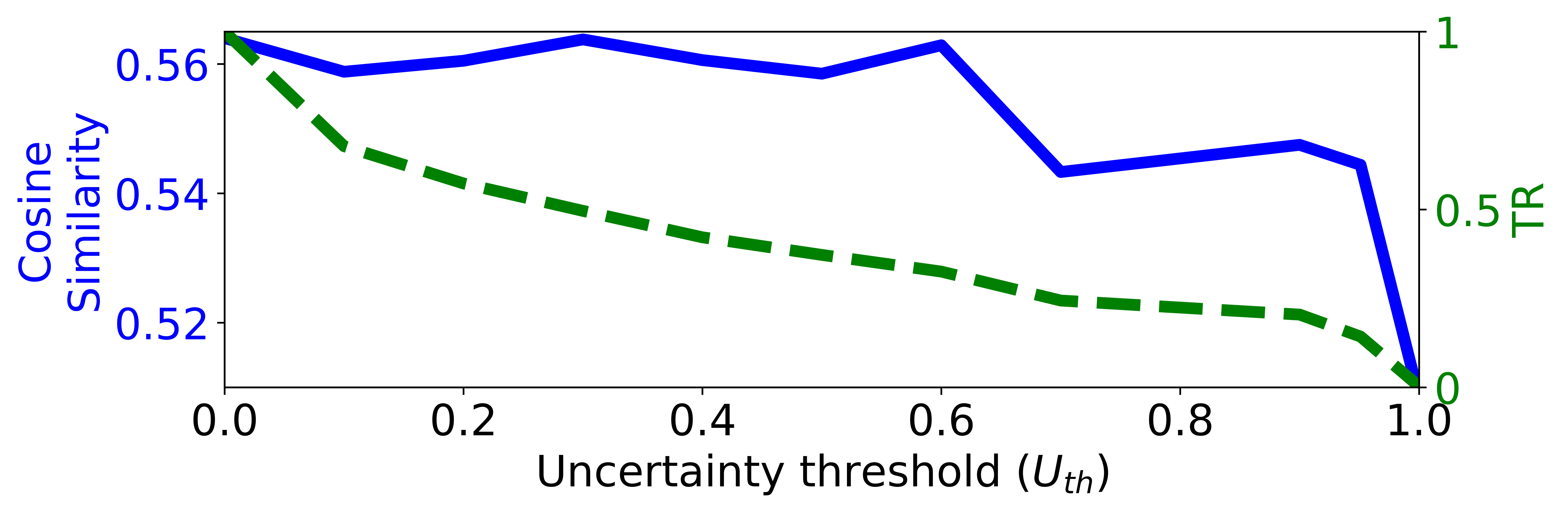}
\vspace{-5pt}
\caption{Cosine similarity and Transmission Rate (TR) of U-HLM as a function of uncertainty threshold.}
\label{fig:cos_sim_tr}
\vspace{-10pt}
\end{figure}

\begin{figure}[t]
\centering \vspace{-5pt}
\includegraphics[width=0.48\textwidth]{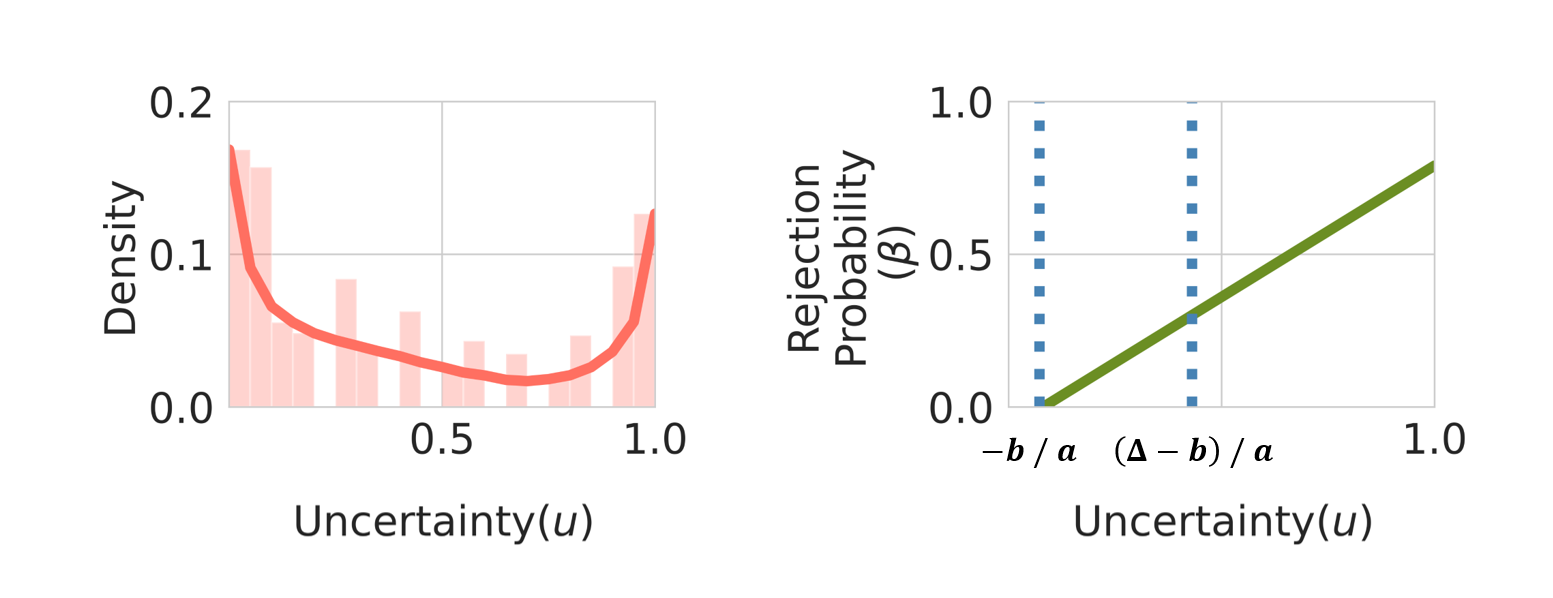}
\vspace{-10pt}
\caption{Empirical probability density of uncertainty (left) and the linear regression curve showing the relationship between uncertainty and rejection probability (right). Dashed vertical lines indicate the two uncertainty thresholds.} 
\vspace{-10pt}
\label{fig:uncertainty_prob_dist}
\end{figure}

\vspace{-5pt}
\section{Numerical Evaluation}
\label{Sec: 5} 
This section demonstrates the performance of the proposed U-HLM. Experiments use TinyLlama-1.1B as the SLM, Llama2-7B as the LLM, and 100 randomly selected samples from the Alpaca dataset \cite{alpaca} as input prompts, conducted on a Linux-based server with an 8-core Intel Xeon Silver 4215R CPU and $3\times$ Nvidia GeForce RTX 3090 GPUs. Additionally, Table \ref{table:generation_performance} includes labeled datasets QED, CREAK, and StrategyQA from the FLAN collection \cite{longpre2023flan}. Baseline methods include LLM, SLM, HLM, and Rand-HLM. Among the HLM methods, speculative inference \cite{leviathan2023fast} is particularly used, offering comparable performance while achieving higher token throughput due to accelerated LLM computation. In the case of Rand-HLM, uplink transmissions are skipped opportunistically based on random probability.

The performance metrics are as follows: inference accuracy is measured using cosine similarity, with the Universal Sentence Encoder (USE) \cite{cer2018universal} employed to transform both the response token sequence and the ground-truth response into vectors of the same dimension; Token Throughput is measured by dividing the total number of generated tokens by the total computation and communication latency; Transmission Rate (TR) measures the percentage of uplink transmission sent when U-HLM or Rand-HLM is applied; True Skip Rate (TSR) indicates the probability of skipping an uplink transmission for a token that would otherwise be accepted. The computation latencies of both the SLM and LLM are measured using wall time. The other hyperparameters used are as follows: $W=1$MHz, $\{p,N\}=\{23,-104\}$dBm, $\alpha=4$, $\rho=2.5$km, and $|\mathcal{V}| = 32,000$. 

\begin{table}[t!]
    \centering
    \caption{Inference accuracy of various inference methods with respect to datasets.} \vspace{-3pt}
        \begin{tabular}{lcccc}
        \toprule
            \multirow{2}{*}{Inference Method} &  \multicolumn{4}{c}{Cosine Similarity}\\ 
            \cmidrule(rrrr){2-5} 
             & Alpaca & QED & CREAK & StrategyQA \\ 
            \midrule
            LLM & 0.6231 & 0.5104 & 0.4862 & 0.6023\\
            SLM & 0.5077 & 0.2577 & 0.2436 & 0.3091\\
            HLM  & 0.5634  & 0.5093 & 0.5040 & 0.5785 \\
            \midrule 
            Rand-HLM & 0.5285 & 0.4638 & 0.4663 & 0.5735\\ 
            \textbf{U-HLM} & 0.5585 & 0.4943 & 0.5130 & 0.5913 \\  
        \midrule
        \end{tabular}
    \label{table:generation_performance}
     \vspace{-15pt}
\end{table}

\vspace{-6pt}
\subsection{Validation for \textbf{Theorem} \ref{theorem:combined}} 
We first calculate the uncertainty thresholds from \textbf{Theorem} \ref{theorem:combined} using the actual measured values. Our experiment provides \(\Delta = 0.301\), with \(a\) and \(b\) already known. This yields \(\frac{\Delta - b}{a} = 0.431\) for risk-prone skipping and \(-\frac{b}{a} = 0.073\) for risk-averse skipping, where the latter is graphically evident in \figref{fig:uncertainty_prob_dist} (right). As shown in \figref{fig:cos_sim_tr}, increasing \(u_{\text{th}}\) generally decreases TR, leading to a reduction in cosine similarity. However, the cosine similarity loss remains negligible up to \(u_{\text{th}} = 0.6\), which corresponds to the experimental uncertainty threshold. Comparing these thresholds, the experimental value lies between those of risk-prone and risk-averse skipping, closer to the former.

Next, we derive the upper bound of \(R\). Instead of using the continuous PDF, we utilize the discrete empirical probability density of uncertainty shown in \figref{fig:uncertainty_prob_dist} (left), yielding \(R < 4\times 10^{-3}\) from \eqref{upperbound}. Experimentally, we observe \(R = 2.24\times 10^{-3}\), confirming that the upper bound is satisfied. This small expected rejection risk $R$ aligns with the minimal cosine similarity loss observed at the uncertainty threshold for risk-prone skipping, as shown in \figref{fig:cos_sim_tr}. This minimal loss can be attributed to the low probability values of uncertainty between the two thresholds, as illustrated in \figref{fig:uncertainty_prob_dist}. Based on these findings, we adopt an uncertainty threshold of \(u_{\text{th}} = 0.431\) for risk-prone skipping.
\vspace{-5pt}
\subsection{Inference Accuracy and Token Throughput Comparisons}
\label{experiment 1 Performance Comparison Across Groups} 
\begin{figure}[t!]
    \centering
    \includegraphics[width=0.99\linewidth]{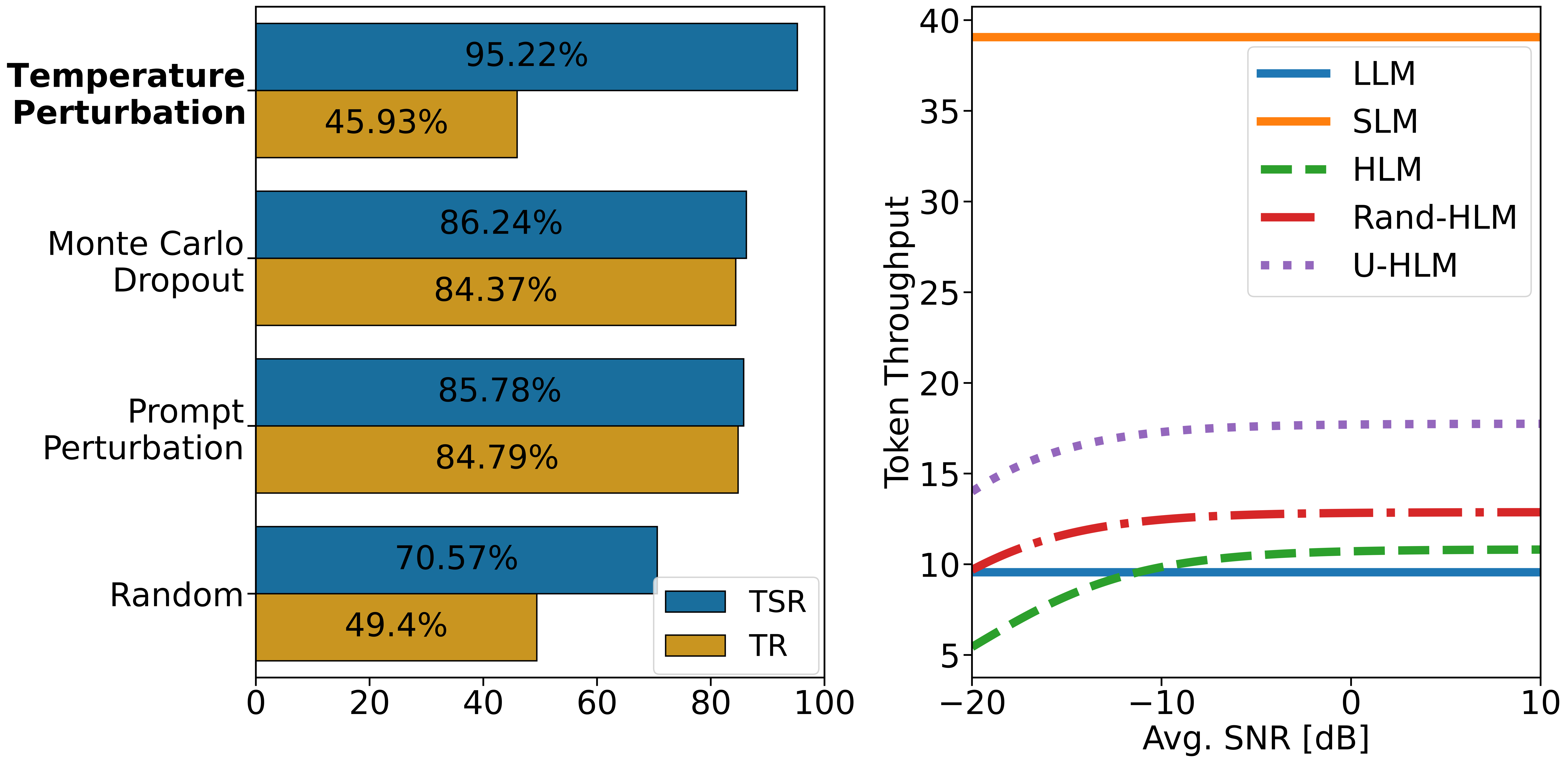}
    \caption{True Skip Rate (TSR) and Transmission Rate (TR) for different HLM inference methods (left); Token throughput versus average SNR for various inference methods (right).}
    \label{fig:MTTvsSNR}
     \vspace{-15pt}
\end{figure}

\vspace{-2pt}
Table \ref{table:generation_performance} and \figref{fig:MTTvsSNR} compare the inference accuracy and token throughput of U-HLM with other inference methods. Table \ref{table:generation_performance} demonstrates that U-HLM outperforms both SLM and Rand-HLM in terms of cosine similarity across all datasets. For instance, with the Alpaca dataset, while Rand-HLM improves cosine similarity by only 2.08\% over SLM, U-HLM achieves a 5.08\% improvement—representing a 244\% greater accuracy gain. On average, U-HLM achieves up to 97.54\% and 100.05\% of the inference accuracy of LLM and HLM, respectively. As shown in \figref{fig:MTTvsSNR} (right), U-HLM delivers the highest token throughput across all SNRs, second only to SLM, with a particularly notable 2.54$\times$ increase in token throughput compared to HLM without skipping.

This advantage stems from U-HLM's ability to accurately and frequently skip uplinks for tokens while maintaining inference quality. This is further evidenced by its higher TSR and lower TR compared to Rand-HLM in \figref{fig:MTTvsSNR} (left), reflecting a 45.93\% reduction in uplink transmissions and LLM computations. Moreover, U-HLM demonstrates substantial improvements in TSR, TR, and token throughput, particularly when temperature perturbation is applied and under low average SNR conditions. This is because, as shown in \figref{fig:uncertainty_comparison}, temperature perturbation facilitates accurate rejection probability estimation due to its linearity, which exhibits a higher slope over a broader uncertainty range. Notably, this increased slope also reduces the upper bound of $R$ in \eqref{upperbound}, potentially enhancing inference accuracy.


\section{Concluding Remarks} 
This paper focused on achieving high-throughput, high-accurate inference in hybrid language models operating over wireless networks. By estimating LLM rejection probabilities through the SLM's uncertainty measurement, the proposed U-HLM enabled on-device opportunistic skipping of uplink transmissions, significantly improving token throughput while preserving inference accuracy. While this work concentrated on token generation, we believe that U-HLM can be extended to other applications, such as token communication, which will be addressed in future work.

\subsubsection*{Acknowledgments}
This work was supported in part by the Korean Ministry of Science and ICT (MSIT) under the National Research Foundation of Korea (NRF) grant (No. 2023-11-1836), in part by Institute of Information \& communications Technology Planning \& Evaluation (IITP) grant funded by MSIT (No. RS-2024-00404972), in part by IITP under YKCS Open RAN Global Collaboration Center (IITP-2024-RS-2024-00434743) grant funded by MSIT, in part by IITP-ITRC (Information Technology Research Center) grant funded by MSIT (IITP-2025-RS-2023-00259991), and in part by SUTD Kickstarter Initiative (SKI 2021\_06\_08).

\bibliographystyle{ieeetr}
\bibliography{main}

\begin{thebibliography}{10}

\bibitem{hoffmann2022training}
J.~Hoffmann~et al., ``Training compute-optimal large language models,'' {\em arXiv preprint arXiv:2203.15556}, 2022.

\bibitem{yang2024harnessing}
J.~Yang~et al., ``Harnessing the power of llms in practice: A survey on chatgpt and beyond,'' {\em ACM KDD}, vol.~18, no.~6, pp.~1--32, 2024.

\bibitem{ma2023llm}
X.~Ma, G.~Fang, and X.~Wang, ``Llm-pruner: On the structural pruning of large language models,'' {\em Advances in neural information processing systems}, vol.~36, pp.~21702--21720, 2023.

\bibitem{li2023loftq}
Y.~Li~et al., ``Loftq: Lora-fine-tuning-aware quantization for large language models,'' {\em arXiv preprint arXiv:2310.08659}, 2023.

\bibitem{zhou2023distillspec}
Y.~Zhou~et al., ``Distillspec: Improving speculative decoding via knowledge distillation,'' {\em arXiv preprint arXiv:2310.08461}, 2023.

\bibitem{hao2024hybrid}
Z.~Hao, H.~Jiang, S.~Jiang, J.~Ren, and T.~Cao, ``Hybrid slm and llm for edge-cloud collaborative inference,'' in {\em Proceedings of the Workshop on Edge and Mobile Foundation Models}, pp.~36--41, 2024.

\bibitem{leviathan2023fast}
Y.~Leviathan, M.~Kalman, and Y.~Matias, ``Fast inference from transformers via speculative decoding,'' in {\em International Conference on Machine Learning}, pp.~19274--19286, PMLR, 2023.

\bibitem{chib1995understanding}
S.~Chib and E.~Greenberg, ``Understanding the metropolis-hastings algorithm,'' {\em The american statistician}, vol.~49, no.~4, pp.~327--335, 1995.

\bibitem{gal2016dropout}
Y.~Gal and Z.~Ghahramani, ``Dropout as a bayesian approximation: Representing model uncertainty in deep learning,'' in {\em international conference on machine learning}, pp.~1050--1059, PMLR, 2016.

\bibitem{huang2023look}
Y.~Huang~et al., ``Look before you leap: An exploratory study of uncertainty measurement for large language models,'' {\em arXiv preprint arXiv:2307.10236}, 2023.

\bibitem{gao2024spuq}
X.~Gao, J.~Zhang, L.~Mouatadid, and K.~Das, ``Spuq: Perturbation-based uncertainty quantification for large language models,'' {\em arXiv preprint arXiv:2403.02509}, 2024.

\bibitem{miller1995wordnet}
G.~A. Miller, ``Wordnet: a lexical database for english,'' {\em Communications of the ACM}, vol.~38, no.~11, pp.~39--41, 1995.

\bibitem{alpaca}
R.~T. et~al., ``Stanford alpaca: An instruction-following llama model,'' 2023.

\bibitem{longpre2023flan}
S.~Longpre~et al., ``The flan collection: Designing data and methods for effective instruction tuning,'' in {\em International Conference on Machine Learning}, pp.~22631--22648, PMLR, 2023.

\bibitem{cer2018universal}
D.~Cer, ``Universal sentence encoder,'' {\em arXiv preprint arXiv:1803.11175}, 2018.

\end{thebibliography}
\end{document}